\pdfoutput=1
\documentclass[11pt]{article}

\usepackage{acl}

\usepackage{times}
\usepackage{latexsym}
\usepackage{graphicx}
\usepackage{multirow}
\usepackage{tabularx}
\usepackage{amsmath}
\usepackage{amssymb}
\usepackage{subcaption}
\usepackage{caption}
\usepackage{booktabs} 
\usepackage{hyperref}
\usepackage[T1]{fontenc}
\usepackage{subcaption}
\usepackage{graphicx}
\usepackage[utf8]{inputenc}
\usepackage{microtype}
\usepackage{float}

\usepackage{inconsolata}

\title{Emotion-Anchored Contrastive Learning Framework for \\ Emotion Recognition in Conversation}

\author{
    Fangxu Yu \quad Junjie Guo \quad Zhen Wu{\thanks{~~~Corresponding author.}} \quad {\bf Xinyu Dai} \\
    National Key Laboratory for Novel Software Technology, Nanjing University, China \\School of Artificial Intelligence, Nanjing University, China \\
    {\tt \{yufx, guojj\}@smail.nju.edu.cn} \\
    {\tt \{wuz, daixinyu\}@nju.edu.cn} \\
}

\begin{document}
\maketitle
\begin{abstract}
Emotion Recognition in Conversation (ERC) involves detecting the underlying emotion behind each utterance within a conversation. Effectively generating representations for utterances remains a significant challenge in this task. Recent works propose various models to address this issue, but they still struggle with differentiating similar emotions such as excitement and happiness. To alleviate this problem, We propose an \textbf{E}motion-\textbf{A}nchored \textbf{C}ontrastive \textbf{L}earning (EACL) framework that can generate more distinguishable utterance representations for similar emotions. To achieve this, we utilize label encodings as anchors to guide the learning of utterance representations and design an auxiliary loss to ensure the effective separation of anchors for similar emotions. Moreover, an additional adaptation process is proposed to adapt anchors to serve as effective classifiers to improve classification performance. Across extensive experiments, our proposed EACL achieves state-of-the-art emotion recognition performance and exhibits superior performance on similar emotions. Our code is available at \href{https://github.com/Yu-Fangxu/EACL}{https://github.com/Yu-Fangxu/EACL}.
\end{abstract}

\section{Introduction}
\label{sec:intro}

\begin{figure}[t]
    \centering
    \includegraphics[width=1.0\columnwidth]{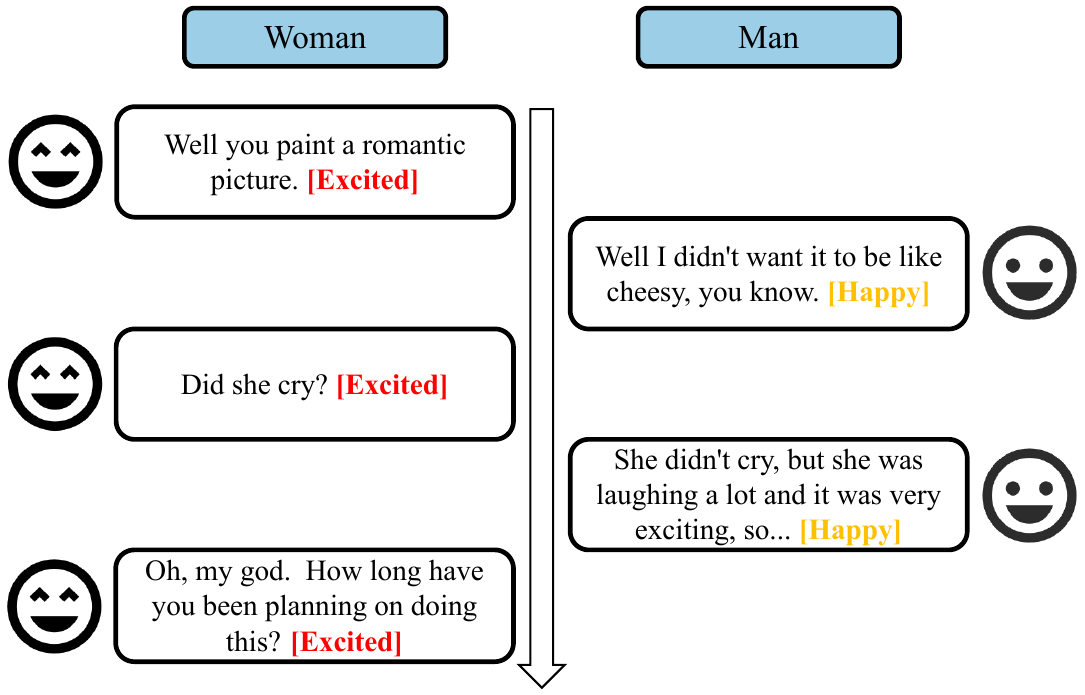}
    \caption{An example of a conversation in the IEMOCAP dataset.}
    \label{fig:conversation}
\end{figure}

Emotion Recognition in Conversation (ERC) aims to identify the emotions of each utterance in a conversation. It plays an important role in various scenarios, such as chatbots, healthcare applications, and opinion mining on social media. However, the ERC task faces several challenges. Depending on the context, similar statements may exhibit entirely different emotional attributes. Simultaneously, distinguishing conversation texts that contain similar emotional attributes is also extremely difficult~\cite{ong2022discourse, zhang2023dualgats}. Figure \ref{fig:conversation} is an example of a chat between a man and a woman. Differentiating between \textit{happy} and \textit{excited} can be challenging for machines due to their frequent occurrence in similar contexts. Appendix \ref{ESA} exhibits quantitative analysis for emotions. This requires the model to accurately distinguish different emotions based on the context. 

\begin{figure}[t]
    \centering
    \includegraphics[width=0.95\columnwidth]{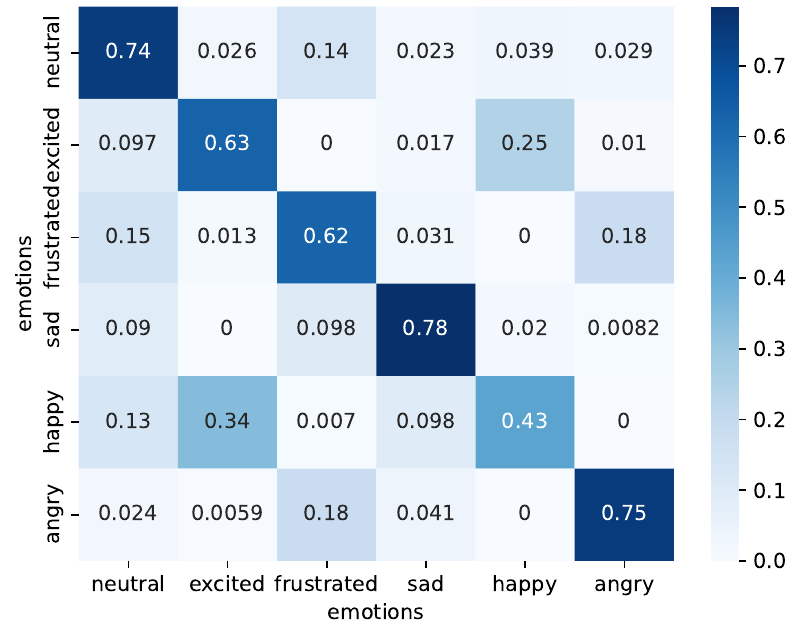}
    \caption{Normalized confusion matrix of SPCL on the IEMOCAP dataset. The rows and columns represent the actual classes and predictions made by the model respectively. The cross-point ($i$, $j$) means the percentage of emotion $i$ predicted to be emotion $j$. Except for the diagonal, the bigger values and deeper color mean these emotions are easily misclassified.}
    \label{fig:wheel}
\end{figure}

Therefore, abundant efforts have been made implicitly to obtain distinguishable utterance representations from two lines, model design and representation learning. As the representative of the former line, DialogueRNN~\cite{majumder2019dialoguernn} designs recurrent modules to track dialogue history for classification. Representation learning methods primarily exploit supervised contrastive learning (SupCon)~\cite{khosla2020supervised} for learning utterance representations. SPCL~\cite{song2022supervised} proposes a prototypical contrastive learning method to alleviate the class imbalance problem and achieve state-of-the-art performance.
Our preliminary fine-grained experimental results for SPCL, as shown in Figure~\ref{fig:wheel}, use the normalized confusion matrix to evaluate the prediction performance. The findings reveal that similar emotions such as \textit{happy} and \textit{excited} are frequently misclassified as each other. This suggests that SPCL still struggles with effectively differentiating similar emotoins.

To tackle the aforementioned issues, this paper presents a novel \textbf{E}motion-\textbf{A}nchored \textbf{C}ontrastive \textbf{L}earning framework (EACL). EACL utilizes textual emotion labels to generate anchors that are emotionally semantic-rich representations. These representations as anchors explicitly strengthen the distinction between similar emotions in the representation space.
Specifically, we introduce a penalty loss that encourages the corresponding emotion anchors to distribute uniformly in the representation space. By doing so, uniformly distributed emotion anchors guide utterance representations with similar emotions to learn larger dissimilarities, leading to enhanced discriminability. After generating separable utterance representations, we aim to compute the optimal positions of emotion anchors to which utterance representations can be assigned for classification purposes. To achieve better assignment, inspired by the two-stage frameworks~\cite{kang2019decoupling, menon2020long, nam2023decoupled}, we propose the second stage to shift the decision boundaries of emotion anchors with fixed utterance representations and achieve better classification performance, which is simple yet effective.

We conduct experiments on three widely used benchmark datasets, the results demonstrate that EACL achieves a new state-of-the-art performance. Moreover, EACL achieves a significantly higher separability in similar emotions, which validates the effectiveness of our method.

The main contributions of this work are summarized as follows:
\begin{itemize}
    \item We propose a novel emotion-anchored contrastive learning framework for ERC, that can generate more distinguishable representations for utterances.
    \item To the best of our knowledge, our method is the first to explicitly alleviate the problem of emotion similarity by introducing label semantic information in modeling for ERC, which can effectively guide representation learning.
    \item Experimental results show that our proposed EACL achieves a new state-of-the-art performance on benchmark datasets. 
\end{itemize}

\section{Related Work}

\subsection{Emotion Recognition in Conversation}
Most of the present works adopt graph-based and sequence-based methods.
DialogueGCN~\cite{ghosal2019dialoguegcn} builds a graph treating utterances as nodes, and models intra-speaker and inter-speaker relationships by setting different edge types between two nodes. MMGCN~\cite{hu2021mmgcn} fuses multi-modal utterance representations into a graph. Differently, DAG-ERC~\cite{shen2021directed} exploits directed acyclic graphs to naturally capture the spatial and temporal structure of the dialogue. COGMEN~\cite{joshi2022cogmen} combines graph neural network and graph transformer to leverage both local and global information respectively. 

Another group of works exploits transformers and recurrent models to learn the interactions between utterances. DialogueRNN~\cite{majumder2019dialoguernn} combines several RNNs to model dialogue dynamics. DialogueCRN~\cite{hu2021dialoguecrn} introduces a cognitive reasoning module. Commensense Knowledge is explored by KET~\cite{zhong2019knowledge} and COSMIC~\cite{ghosal2020cosmic}. Cog-BART~\cite{li2022contrast} employs BART~\cite{lewis2019bart} to simultaneously generate responses and detect emotions with the auxiliary of contrastive learning. EmoCaps~\cite{li2022emocaps} and DialogueEIN~\cite{liu2022dialogueein} design several modules to explicitly model emotional tendency and inertia, local and global information in dialogue. The power of the language models is utilized by CoMPM~\cite{lee2021compm} which learns and tracks contextual information by the language model itself and SPCL~\cite{song2022supervised}, a prototypical supervised contrastive learning method to alleviate the data imbalance problem. SACL~\cite{hu2023supervised}introduces adversarial examples to learn robust representations. Our EACL goes along this track. Unlike the above approaches, HCL~\cite{yang2022hybrid} comes up with a general curriculum learning paradigm that can be applied to all ERC models. InstructERC~\cite{lei2023instructerc} and DialogueLLM~\cite{zhang2023dialoguellm} construct instructions and fine-tune LLMs for ERC. ~\cite{lee2022emotion, guo2021label} learn from soft labels.
\subsection{Supervised Contrastive Learning}
Recent works~\cite{chen2020simple,he2020momentum} in unsupervised contrastive learning provide a similarity-based learning framework for representation learning. These methods maximize the similarity between positive samples while minimizing the similarity between negative sample pairs. To make use of supervised information, supervised contrastive learning (SupCon)~\cite{gunel2020supervised} aims to make the data that have the same label closer in the representation space and push away those that have different labels. However, SupCon works poorly in data imbalance settings. To mitigate this problem, KCL~\cite{kang2021exploring} explicitly pursues a balanced representation space. TSC~\cite{li2022targeted} uniformly set targets in the hypersphere and enforce data representations to close to the targets. BCL~\cite{zhu2022balanced} regards classifier weights as prototypes in the representation space and incorporates them in the contrastive loss. LaCon~\cite{zhang2022label} incorporates label embedding for better language understanding. Our method is inspired by TSC, differently, we incorporate emotion semantics in the representation space and dynamically adjust the emotion anchors for better classification.
\section{Methodology}
\subsection{Problem Definition}
A conversation can be denoted as a sequence of utterances $\{u_1,u_2,u_3,...,u_n\}$, each utterance $u_t$ is uttered by one of the conversation speakers $s_j$. There are $m\quad(m\geq 2)$ speakers in the conversation, denoted as $\{s_1,s_2,...,s_m\}$. Given the set of emotion labels $ \mathcal{E} $ and conversation context $\{(u_1, s_{u_1}),(u_2, s_{u_2}),...,(u_t, s_{u_t})\}$, the ERC task aims to predict emotion $e_t(e_t \in \mathcal{E})$ for current utterance $u_t$. $\mathcal{E}$ is a set of emotions. For instance, in the IEMOCAP dataset, $\mathcal{E}$ = \{\textit{excited}, \textit{frustrated}, \textit{sad}, \textit{neutral}, \textit{angry}, \textit{happy}\}.

\begin{figure*}[h]   
	
	\centering
	
	\includegraphics[width=2.05\columnwidth]{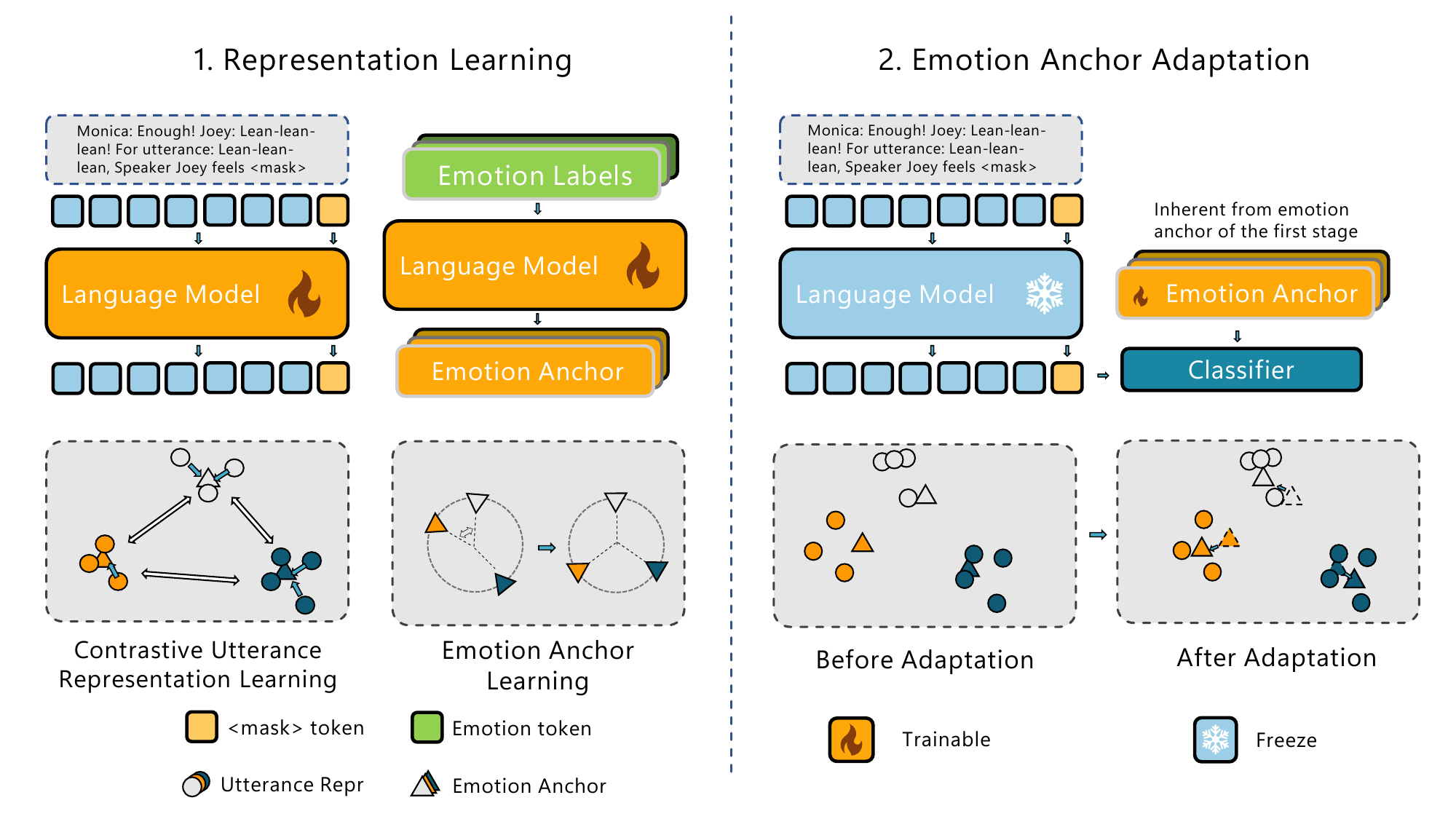}
	
	\caption{Overview of our proposed framework. \textbf{Left side} introduces representation learning, which is composed of utterance representation and emotion anchor learning. \textbf{Right side} describes the process of adapting emotion anchors to the optimal positions for classification.}
	
	\label{fig:main_arch}
	
\end{figure*}

\subsection{Model Overview}

The overview of our model is shown in Figure \ref{fig:main_arch}. The encoding strategy of our model adopts the paradigm of prompt learning (Section \ref{sec:cen}). Our training process is composed of two stages. 

The first stage (Section \ref{sec:stage1}) is called representation learning, which aims to learn more distinctive representations with emotion anchors. Concretely, we incorporate anchors containing semantic information into the contrastive learning framework and utilize them to guide the learning of utterance representations. Our objectives are (1) to bring utterances with the same emotion closer to their corresponding anchors and push utterances with different emotions farther away, and (2) to achieve a more uniform distribution of anchors in the hyperspace for better classifying different emotions.

The second stage (Section \ref{sec:stage2}) is called emotion anchor adaptation, which aims to further improve classification performance by slightly adjusting anchors. The anchors in the first stage can help the model learn separable representations of utterances. However, separated emotion anchors may not be located in the most representative positions of each category of utterance representation for the following emotion recognition because contrastive learning in the first stage aims not to achieve this goal. Therefore, we design the second stage to slightly adjust the positions of emotion anchors to shift the decision boundaries for better classification performance. In this stage, we freeze the parameters of the language model and only fine-tune the emotion anchors, as shown on the right side of Figure \ref{fig:main_arch}. 
Lastly, EACL matches the utterance representations with the most similar emotion anchors to make predictions.

\subsection{Prompt Context Encoding}
\label{sec:cen}
Following previous work~\cite{song2022supervised}, we employ pre-trained language models and adopt prompt tuning to transform the classification into masked language modeling. An effective prompt template aligns the downstream task with the large semantic information learned by the language model in the pre-training stage, which boosts the model's performance in downstream tasks.

To predict the emotion of utterance $u_t$, we take $k$ utterances before timestamp $t$ as the context to predict $e_t$. Formally, the input for the language model is composed as: 
\begin{equation}
    x_t = [s_{t-k}, u_{t-k}, \ldots, s_{t}, u_t, Prompt]
\end{equation}
where Prompt $P$ = "For utterance $u_t$, speaker $s_t$ feels [mask]" . We take the last hidden state of [mask] as utterance representation.

\subsection{Stage One: Representation Learning}\label{URL}
\label{sec:stage1}
In this section, we will introduce two main components of EACL in stage one: utterance representation learning and emotion anchor learning. 
\subsubsection{Utterance Representation Learning}
\label{sec:url}

The objective in this section is to acquire discernible representations for each individual utterance. To accomplish this, we employ label encodings to generate emotion anchors and incorporate them into a contrastive learning framework. By utilizing these anchors, we can proficiently steer the process of representation learning.

Given a batch of samples $\mathcal{X} = \{ x_1, x_2, \ldots, x_b \} \\ \in \mathbb{R}^{b \times \ell}$, where $b, \ell$ are batch size and max length of input respectively.
We feed $\mathcal{X}$ into the pre-trained language model and get the last hidden states $\mathcal{Z} = {\rm Encoder}(\mathcal{X})$. Then we use the hidden state of [mask] token at the end of the sentence as the representation of utterance $u_t$. Finally, we obtain the representations of utterances with an MLP layer:
\begin{align}
    \mathcal{R} = {\rm MLP}_{cl}(\mathcal{Z}_{[mask]})
\end{align}
where $\mathcal{R}=\{ r_1, r_2, \ldots, r_b \}$ and $ \mathcal{R} \in \mathbb{R}^{b\times d}$, $d$ is dimension of the encoder.

Similarly, we take textual emotion labels as the input of language models to obtain emotion anchors for all emotions $\mathcal{E} = \{e_1,e_2,\ldots,e_s\}$:
\begin{equation}
\begin{aligned}
    \mathcal{Z}^{a} & = {\rm Encoder}(\mathcal{E}) \\
    \mathcal{A} &= {\rm MLP}_{cl}(\mathcal{Z}^a)
\end{aligned}    
\end{equation}
where $\mathcal{A} \in \mathbb{R}^{s \times d}$, each row of which represents a emotion anchor. $s$ represents the number of emotions. To ensure we get a stable anchor representation, $\mathcal{Z}_a$ is frozen in our training process.

We propose an emotion-anchored contrastive learning loss to utilize emotion label semantics for better representation learning. More specifically, in each mini-batch, we let $\mathcal{V} = \{v_1,v_2,\ldots,v_{b+s}\} = \mathcal{R} \cup \mathcal{A}$ and $\mathcal{V}^+_i$ represents the set of utterances or anchor representation that have the same label as utterance $r_i$ except for itself. Finally, our emotion-anchored contrastive loss is as follows:
\begin{equation}
    \begin{aligned}
    c_{ij} &= {\rm sim}(v_i, v_j) / \tau
    \\
    \mathcal{L}_{sup} &= \sum_{i=1}^{s + b}-\log \frac{\sum_{v_j \in \mathcal{V}^+_i}{e^{c_{ij}}}}{|\mathcal{V}^+_i|\sum_{v_j \in \mathcal{V}} e^{c_{ij}} }
     \end{aligned}
\label{escl}
\end{equation}
where $|\mathcal{V}^+_i|$ represents number of positive examples. $\tau$ is the temperature hyperparameter for the contrastive loss. ${\rm sim}$ represents a similarity function, we adopt cosine similarity here.

In equation~\ref{escl}, the interactions between representations can be divided into three components: utterances-utterances, anchors-utterances, and anchors-anchors. Representations with the same label are brought closer to each other, while those with different labels are pushed farther apart. The utterances-utterances interactions are similar to traditional contrastive learning, while the anchors-utterances interactions represent the process of anchor-guided utterance representation learning. The anchors-anchors interaction ensures a better distinction between different emotions. 

Recent research~\cite{gunel2020supervised} has indicated that combining cross-entropy loss with contrastive learning facilitates language models with more discriminative ability. Therefore cross-entropy loss is added to help improve representation learning. We additionally add a linear mapping for classification:
\begin{equation}
    \hat{\mathcal{Y}} = {\rm softmax}({\rm MLP_{ce}}(\mathcal{Z}_{[mask]}))
\end{equation}
\begin{equation}
    \mathcal{L}_{CE} = -\frac{1}{b}\sum_{i=1}^{b}\sum_{j=1}^{s}y_{ij}\log \hat{y}_{ij}
\end{equation}
where $\hat{\mathcal{Y}} \in \mathbb{R}^{b \times s}$ represents the possibility distribution of $b$ utterances over $s$ emotions. $y_{ij}$ represents the element in the $i$-th row and $j$-th column of $\hat{\mathcal{Y}}$. ${\rm MLP}_{ce}$ is a linear layer for classification. 

\subsubsection{Emotion Anchor Learning}
 Nevertheless, despite the implementation of the interaction between representations, the three types of interactions mentioned in Section \ref{sec:url} alone are insufficient to explicitly disperse the distance between the most similar emotion anchors. To further tackle the issue of similarity, we propose an anchor angle loss. This loss is designed to incentivize emotion anchors to maximize the angle between themselves and their most similar emotion anchors within the contrastive space:

\begin{equation}
    \mathcal{L}_{Ag} = - \frac{1}{s}\sum_{i=1}^{s} \min_{j, i \neq j}\arccos \frac{\left \langle a_i, a_j \right \rangle }{\| a_i\| \| a_j\|} 
\end{equation}
where $a_i$ represents $i$-th emotion anchor representation in $\mathcal{A}$.

$\mathcal{L}_{Ag}$ aims to minimize the maximal pairwise cosine similarity between all the emotion anchors. It is equivalent to maximizing the minimal pairwise angle. The more dispersed emotion anchors are, the better their capacity to recognize similar emotions.

Combining all the components mentioned in stage one, the overall loss is a weighted average of cross-entropy loss, anchor angle loss, and contrastive loss, as given in equation \ref{total loss}.
\begin{equation}
    \mathcal{L} = \lambda_1 (\mathcal{L}_{sup} + \lambda_2 \mathcal{L}_{Ag}) + (1 - \lambda_1)\mathcal{L}_{CE} 
\label{total loss}
\end{equation}
where $\lambda_1$ and $\lambda_2$ are hyper-parameters to balance loss terms.

\subsection{Stage Two: Emotion Anchor Adaptation}
\label{sec:stage2}
In the first stage, we used emotion anchors generated from emotion labels to guide the convergence of utterance representations toward different emotion clusters. These emotion anchors serve as representatives for each emotion, which are suitable to function as effective nearest-neighbor classifiers for utterance representations. However, separated emotion anchors trained from stage one may not be located in the most representative positions of each category of utterance representation, which weakens the classification ability of emotion anchors. To ensure the alignment between utterance representations and emotion anchors,
we propose the second stage to adapt the emotion anchors to shift the decision boundaries by training them with a small number of epochs. This approach aims to enhance the ability of emotion anchors for classification purposes.
 
To be more specific, we freeze the parameters of the language model and make the emotion anchors inherited from stage one $a_i(i=1,...,s)$ trainable parameters, which corresponds to the right side in Figure \ref{fig:main_arch}. In order to be consistent with the representation learning, we still use the same similarity measure for adapting emotion anchors.

The loss function for emotion anchor adaptation:
\begin{equation}
    \begin{aligned}
        c_{ij} &= {\rm sim}(r_i, a_j) / \tau \\
        \mathcal{L}_{ada} 
        &= - \frac{1}{b}\sum_{i=1}^{b}\sum_{j=1}^{s}y_{ij}\log \hat{y_{ij}}\\
        & =  - \frac{1}{b}\sum_{i=1}^{b}\sum_{j=1}^{s}y_{ij}\log \frac{e^{c_{ij}}}{\sum_{k=1}^{s}{e^{c_{ik}}}}
    \end{aligned}
\end{equation}
where $c_{ij}$ means adjusted cosine similarity between the $i$-th utterance representation $r_i$ and $j$-th emotion anchors $a_j$. $\tau$ is the same temperature hyper-parameter in stage one.

\subsection{Emotion Prediction}
During the inference stage, we predict emotion labels by matching each utterance representation with the nearest emotion anchor:
\begin{equation}
    \hat{y}_i = \arg \max_{j} {\rm sim}(r_i, a_j)
\end{equation}
Where $r_i$ is the representation of utterance $x_i$ and $a_j$ is the emotion anchor of class $j$.

\section{Experiments}
\subsection{Experimental setup}
The language model loads the initial parameter with SimCSE-Roberta-Large~\cite{gao2021simcse}. All experiments are conducted on a single NVIDIA A100 GPU 80GB and we implement models with PyTorch 2.0 framework. More experimental details are provided in Appendix \ref{sec:hyper}.

\subsection{Datasets}

In this section, we will introduce three adopted popular benchmark datasets: 
IEMOCAP~\cite{busso2008iemocap}, MELD~\cite{poria2018meld} and EmoryNLP~\cite{zahiri2017emotion}.

(1) \textbf{IEMOCAP}: consists of 151 videos of two speakers' dialogues with 7433 utterances. Each utterance is annotated by an emotion label from 6 classes, including \textit{excited}, \textit{frustrated}, \textit{sad}, \textit{neutral}, \textit{angry}, and \textit{happy}.

(2) \textbf{MELD}: is extracted from the TV show Friends. It contains about 13000 utterances from 1433 dialogues. Each utterance is labeled by one of the following 7 emotion labels: \textit{surprise}, \textit{neutral}, \textit{anger}, \textit{sadness}, \textit{disgusting}, \textit{joy}, and \textit{fear}.

(3) \textbf{EmoryNLP}: contains 97 episodes, 897 scenes, and 12606 utterances from TV show Friends. It differs from MELD in that the emotional tags contained are: \textit{joyful}, \textit{sad}, \textit{powerful}, \textit{mad}, \textit{neutral}, \textit{scared}, and \textit{peaceful}.

In our experiments, we only use textual modality. The detailed statistics of the three datasets are shown in Table \ref{table1}.

\subsection{Metrics}
Following previous works~\cite{lee2021compm, song2022supervised}, we choose the weighted-average F1 score as the evaluation metric.

\begin{table}[H]
\resizebox{\columnwidth}{!}{%
\begin{tabular}{c|cccccc|c}
\toprule
\multirow{2}{*}{Dataset} & \multicolumn{3}{c}{Dialogues} & \multicolumn{3}{c|}{Utterances} & \multirow{2}{*}{CLS} \\ 
         & train & dev & test & train & dev  & test &   \\ \midrule
IEMOCAP  & 100   & 20  & 31   & 4810  & 1000 & 1623 & 6 \\ 
MELD     & 1038  & 114 & 280  & 9989  & 1109 & 2610 & 7 \\ 
EmoryNLP & 659   & 89  & 79   & 7551  & 954  & 984  & 7 \\ \bottomrule
\end{tabular}%
}
\caption{Statistics of the three datasets, where CLS is the number of classes.}
\label{table1}
\end{table}
\subsection{Baselines}
For a comprehensive evaluation, we compare our method with the following baselines:
\begin{table*}[t]
\centering
\resizebox{1.45\columnwidth}{!}{%
\begin{tabular}{ccccc}
\toprule
Methods     & IEMOCAP & MELD  & EmoryNLP & Average\\ \midrule
\multicolumn{5}{c}{\textit{Graph-based models}} \\\midrule
DialogueGCN~\cite{ghosal2019dialoguegcn}  & 64.91   & 63.02 & 38.10  &  55.34  \\ 
RGAT~\cite{ishiwatari2020relation}    & 66.36   & 62.80 & 37.89  & 55.68 \\ 
DAG-ERC~\cite{shen2021directed}     & 68.03   & 63.65 & 39.02 & 56.9  \\ 
DAG-ERC+HCL~\cite{yang2022hybrid} & 68.73   & 63.89 & 39.82  & 57.48 \\
SIGAT~\cite{jia2023speaker} & \underline{70.17}    & 66.20 & 39.95 & 58.77 \\\midrule 
\multicolumn{5}{c}{\textit{ Sequence-based models}} \\ \midrule
COSMIC~\cite{ghosal2020cosmic}       & 65.25   & 65.21 & 38.11 & 56.19   \\ 
+CKCL~\cite{tu2023context}          & 67.16    & 66.21 & \underline{40.23}  & 57.87\\
Cog-BART~\cite{li2022contrast}    & 66.18   & 64.81 & 39.04  & 56.68 \\ 
DialogueEIN~\cite{liu2022dialogueein}  & 68.93   & 65.37 & 38.92  & 57.74 \\ 
CoMPM~\cite{lee2021compm}      & 69.46   & \underline{66.52} & 38.93  & 58.3 \\ 
SupCon~\cite{gunel2020supervised}      & 68.14   & 65.63 & 39.28  & 57.68 \\ 
Emocaps~\cite{li2022emocaps}    & 69.49   & 63.51 & -   & -     \\
SPCL+CL~\cite{song2022supervised} & 67.19   & 65.74 & 39.52  & 57.48 \\ 
SACL~\cite{hu2023supervised}       &   69.22     &    66.45      &   39.65    & \underline{58.44} \\
SCCL~\cite{yang2023cluster} & 69.88    & 65.70 & 38.75 & 58.11 \\
DIEU~\cite{zhao2023dieu}    & 69.90    & 66.43 & 40.12 & 58.81 \\
MPLP~\cite{zhang2023mimicking} & 66.65 & 66.51 & - & - \\
ChatGPT 3-shot \cite{zhao2023chatgpt} & 48.58       &58.35    & 35.92    & 47.62 \\ \hline
EACL (ours)    & \textbf{70.41}      & \textbf{67.12} \textsuperscript{\dag} & \textbf{40.24} & \textbf{59.26}\textsuperscript{\dag}  \\ 
\bottomrule

\end{tabular}%
}
\caption{Weighted-average F1 score of different models on benchmark datasets. Bold font and underlining indicate the best and second-best performance respectively. SPCL+CL is reproduced with the official code and uses the SimCSE-Roberta-Large that EACL uses. \textsuperscript{\dag} represents statistical significantly over baselines with t-test (p<0.05)}
\label{performance}
\end{table*}

(1) Graph-based model:
\textbf{DialogueGCN~\cite{ghosal2019dialoguegcn}} employs GCNs to gather context features for learning utterance representations, Shen~\cite{shen2021directed} shows the performance of replacing the feature extractor with Roberta-Large. 
\textbf{RGAT~\cite{ishiwatari2020relation}} proposes relational position encodings to model both speaker relationship and sequential information. 
\textbf{DAG-ERC~\cite{shen2021directed}} utilizes an acyclic graph neural network to intuitively model a conversation's natural structure without introducing any external information. 
\textbf{DAG-ERC+HCL~\cite{yang2022hybrid}} proposes a curriculum learning paradigm combined with DAG-ERC for learning from easy to hard.
\textbf{SIGAT~\cite{jia2023speaker}} models speaker and sequence information in a unified graph to learn the interactive influence between them. 

\indent (2) Sequence-based model: 
\textbf{COSMIC~\cite{ghosal2020cosmic}}
 incorporates different elements of commonsense and leverages them to learn self-speaker dependency.
\textbf{Cog-BART~\cite{li2022contrast}} applies BART with contrastive learning to take response generation into consideration. 
\textbf{DialogueEIN~\cite{liu2022dialogueein}} designs emotion interaction and tendency blocks to explicitly simulate emotion inertia and stimulus.
\textbf{CoMPM~\cite{lee2021compm}} utilizes pretrained models to directly learn contextual information and track dialogue history.
\textbf{SupCon~\cite{gunel2020supervised}} is the vanilla supervised contrastive learning.
\textbf{SCCL~\cite{yang2023cluster}} conducts contrastive learning with 3-dimensional affect representations. 
\textbf{DIEU~\cite{zhao2023dieu}} aims to solve the long-range context propagation problem.
\textbf{CKCL~\cite{tu2023context}} denoises information irrelevant context and knowledge when training.
\textbf{MPLP~\cite{zhang2023mimicking}} models the history and experience of speakers and exploits paraphrasing to enlarge the difference between labels.
\textbf{Emocaps~\cite{li2022emocaps}} devises transformer to a novel architecture, Emoformer, to extract the emotional tendency of utterance. 
\textbf{SACL~\cite{hu2023supervised}} proposes contrastive learning combined with adversarial training for robust representations.
\textbf{SPCL+CL~\cite{song2022supervised}} combines prototypical contrastive learning and curriculum learning to tackle the emotional class imbalance issue. 
\textbf{ChatGPT~\cite{zhao2023chatgpt}} reports results in the 3-shot performance.

\section{Results and Analysis}
\subsection{Main Results}
\begin{table}[t]

\begin{subtable}[h]{\linewidth}
    \caption{IEMOCAP}
    \resizebox{\columnwidth}{!}{%
        \centering
       \begin{tabular}{ccccccccc}
\toprule
Methods       & Exc & Fru  & Sad & Neu & Ang & Hap & Avg & W-f1\\ \midrule
SPCL+CL &    66.72  & 63.96  & 80.03  & 72.29  & 64.82  & 43.96  &  65.30  & 67.19\\ 
EACL          & \textbf{71.27} & \textbf{67.76} & \textbf{81.80} & \textbf{73.32} & \textbf{67.54} & \textbf{51.29}& \textbf{68.81} & \textbf{70.41}\\ 
$\Delta$  & +4.55 & +3.80 & +1.77 & +1.03 & +2.72 & +7.33 & +3.51 & +3.22\\
\bottomrule
\end{tabular}%
}
\end{subtable}

\begin{subtable}[h]{\linewidth}
\caption{MELD}
\resizebox{\columnwidth}{!}{%
    \centering
   \begin{tabular}{cccccccccc}
\toprule
Methods       & Fear & Neu  & Ang & Sad & Dis & Surp & Joy & Avg& W-f1\\ \midrule
SPCL+CL    & \textbf{26.59}  & 77.92  &  \textbf{54.40}  &  \textbf{43.53} & 30.94 &  59.26 & 60.34  & 50.43  &  65.74 \\ 
EACL          & 23.54  &  \textbf{80.44} & 54.01  &  42.41 & \textbf{33.86}  & \textbf{60.48}  & \textbf{65.22}  & \textbf{51.42} & \textbf{67.12}\\
$\Delta$  & -3.05 & +2.52 & -0.39 & -1.12 & +2.92 & +1.22 & +4.88 & +0.99 & +1.38 
\\ \bottomrule
\end{tabular}%
}
\end{subtable}
\begin{subtable}[h]{\linewidth}
\caption{EmoryNLP}
\resizebox{\columnwidth}{!}{%
\centering
\begin{tabular}{cccccccccc}
\toprule
Methods       & Joy & Sad  & Pow & Mad & Neu & Pea & Sca & Avg & W-f1\\ \midrule
SPCL+CL       & 53.52 & \textbf{31.61}  & 10.28 & \textbf{44.21} & \textbf{51.40} & 16.83& 39.51 & 35.34 & 39.52\\ 
EACL          & \textbf{52.73} & 30.77 & \textbf{15.27} &  41.97 & 49.76  & \textbf{23.48} & \textbf{41.18} & \textbf{36.45} & \textbf{40.24}\\ 
$\Delta$  & -0.79 & -0.84 & +4.99 & -2.24 & -1.64 & +6.65 & +1.67 & +1.11 & +0.72 \\
\bottomrule
\end{tabular}%  
}

\end{subtable}
\caption{Fine-grained performance comparison between SPCL+CL and EACL for all emotions on three benchmark datasets, the F1-score is used for each class. $\Delta$ is the difference between the two models.}
\label{fine}
\end{table}
Table \ref{performance} reports the results of our method and the baselines. Our model outperforms other baselines and achieves a new state-of-the-art performance on IEMOCAP, MELD, and EmoryNLP datasets. The results exhibit the effectiveness of our emotion-anchored contrastive learning framework.  

Based on the results, we can observe that sequence-based methods have overall better performance than graph-based methods. Compared to the graph-based models, EACL improves a large margin over the DAG-ERC~\cite{shen2021directed} which is the state-of-the-art graph-based method without introducing extra knowledge by 2.38\%, 3.57\%, and 1.22\% on three benchmark datasets. 

Compared to sequence-based methods, EACL outperforms two contrastive learning methods, SACL and SPCL+CL by a large margin. Specifically, SPCL's use of a queue for storing class representations and prototype generation from small batches results in unstable representation learning. Significant movement of prototypes that undergo during training and the asynchronous update of queue representations with the language model's parameters lead to suboptimal utterance representations. EACL outperforms the state-of-the-art results on the IEMOCAP dataset by 0.92\%, the MELD dataset by 0.6\%, and the EmoryNLP dataset by 0.59\%. Besides, EACL has an overwhelming performance advantage over ChatGPT, one possible reason is that the few-shot prompt setting may not be enough to achieve satisfactory performance.

Table \ref{fine} reports the fine-grained performance on benchmark datasets. EACL outperforms SPCL+CL which is the most relevant method to us in most emotion categories on all benchmark datasets. Specifically, in the IEMOCAP dataset, We have observed a significant improvement in performance on two pairs of similar emotions, \textit{happy} and \textit{excited} with an increase of 7.33\% and 4.55\%, \textit{frustrated} and \textit{angry} with an increase of 3.80\% and 2.72\% respectively. Detailed performance analysis is provided in Appendix \ref{DPA}.

\begin{table}[t]
\resizebox{\columnwidth}{!}{%
\begin{tabular}{cccc}
\toprule
Dataset       & IEMOCAP & MELD  & EmoryNLP \\ \midrule
Original      & 70.41   & 67.12 & 40.24    \\  \hline
w/o Emotion Anchor Learning & 69.78 (0.63 $\downarrow$)   & 66.63(0.49 $\downarrow$) & 39.90(0.34 $\downarrow$)    \\ 
%- semantic  & 69.41        & 66.99      &  39.42        \\ 
w/o Classification Objective    & 69.98(0.43 $\downarrow$)   & 66.24(0.88 $\downarrow$) & 39.73(0.51 $\downarrow$)    \\  \hline
w/o Anchor Inheritance    & 69.79(0.62 $\downarrow$)   & 67.03(0.09 $\downarrow$) & 38.46 (1.78 $\downarrow$)   \\  
w/o Anchor Adaptation    & 69.67(0.74 $\downarrow$)   & 64.43(2.89 $\downarrow$)     & 39.98 (0.26 $\downarrow$)   \\  
w/  representation center         & 69.84(0.57 $\downarrow$) & 66.49(0.63 $\downarrow$) & 39.84(0.38 $\downarrow$)\\
\bottomrule
\end{tabular}%
}
\caption{Ablation results on benchmark datasets.}
\label{ablation}
\end{table}

\subsection{Ablation Study}
We conduct a series of experiments to confirm the effectiveness of components in our method. The results are shown in Table \ref{ablation}. Removing any element of EACL makes the overall performance worse.

To validate the effects of components in the first stage, We remove the $\mathcal{L}_{Ag}$ which encourages the angle of different emotion anchors to be uniform. We can find that the lack of $\mathcal{L}_{Ag}$ results in a significant decline in the performance of nearly 0.5\%, as reported in line 2 in Table \ref{ablation}, indicating that emotion anchor learning helps for separating utterance representations. Also, the removal of $\mathcal{L}_{CE}$ drops the performance by about 0.5\% on average, the result demonstrates that supervised learning benefits the fine-tuning of language models. 

In the second stage, We explore whether adapting emotion anchors and emotion semantics are necessary. Similar to classifier re-training~\cite{kang2019decoupling, nam2023decoupled}, we randomly initialize emotion anchors that lie far from the data distribution after learning the utterance representations. Training from scratch is a cold start and cannot reach the optimal position. This result in Line 4 verifies the importance of inheriting emotion anchors and the result shows that the trained emotion anchors express a more powerful ability of recognition. When we remove the anchor adaptation or take the center of training representations for each emotion category as emotion anchors, performance will degrade significantly, indicating the improper positions of emotion anchors weaken the classification performance and verifying the importance of stage two. Lines 5 and 6 in Table \ref{ablation} confirms our assumption. In summary, the components of our method contribute to the results substantially.

\begin{table}[t]
\resizebox{\columnwidth}{!}{%
\begin{tabular}{cccc}
\toprule
Dataset       & IEMOCAP & MELD  & EmoryNLP \\ \midrule
SimCSE-Roberta-Large      & 70.41   & 67.12 & 40.24    \\ 
Deberta-Large           & 69.09   & \textbf{67.80} & \textbf{41.09} \\ 
Promcse-Roberta-Large   &  \textbf{70.45}     &  67.38      &   40.93      \\  \bottomrule
\end{tabular}%
}
\caption{Performance under different language models.}
\label{different}
\end{table}

\subsection{Performance on Different Language Models}
To evaluate the versatility of our learning framework, we conducted experiments using different pretrained language models. Specifically, we examined the performance of our framework on two additional popular language models, namely Deberta-Large~\cite{he2020deberta} and Promcse-Roberta-Large~\cite{jiang2022improved}. The results, presented in Table \ref{different}, demonstrate that all the pretrained models deliver competitive performance. This observation serves as evidence for the robustness and effectiveness of our framework across various pre-trained language models. It further emphasizes the generalizability of our approach in conversational emotion recognition tasks. We report fine-grained performance in Appendix~\ref{sec:fine-diff}.

\subsection{Emotion Similarity Comparison}
\label{ESC}
\begin{figure}[t]
\centering
\begin{subfigure}{0.49\columnwidth}
\centering
    \includegraphics[width=1.0\columnwidth]{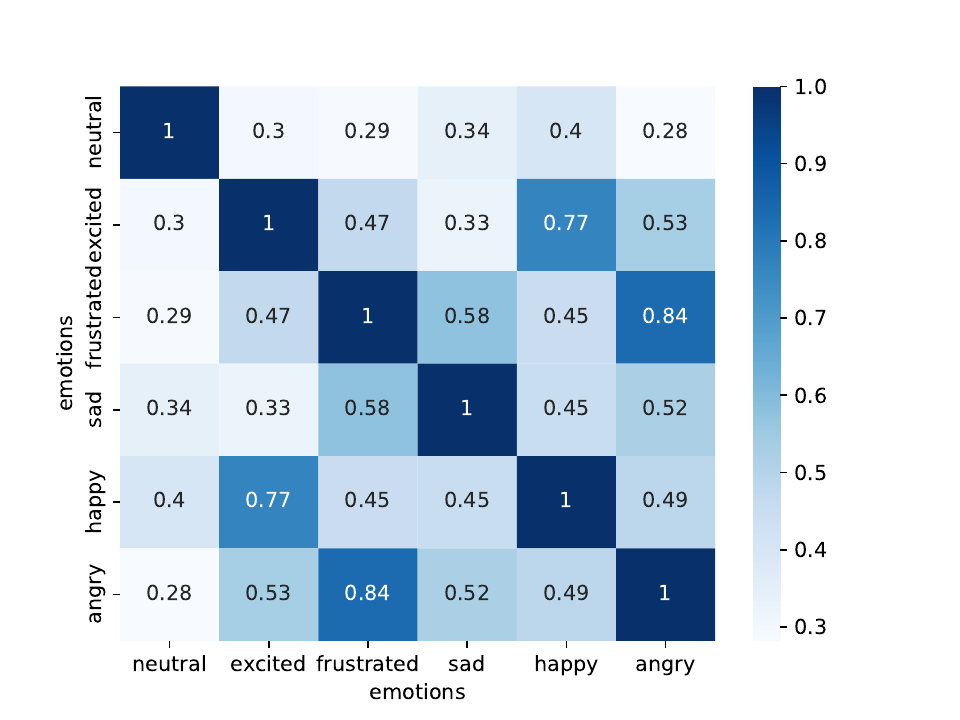}
    \caption{Before training}
    \label{fig:before}
\end{subfigure}
\hfill
\begin{subfigure}{0.49\columnwidth}
    \centering
    \includegraphics[width=1.0\columnwidth]{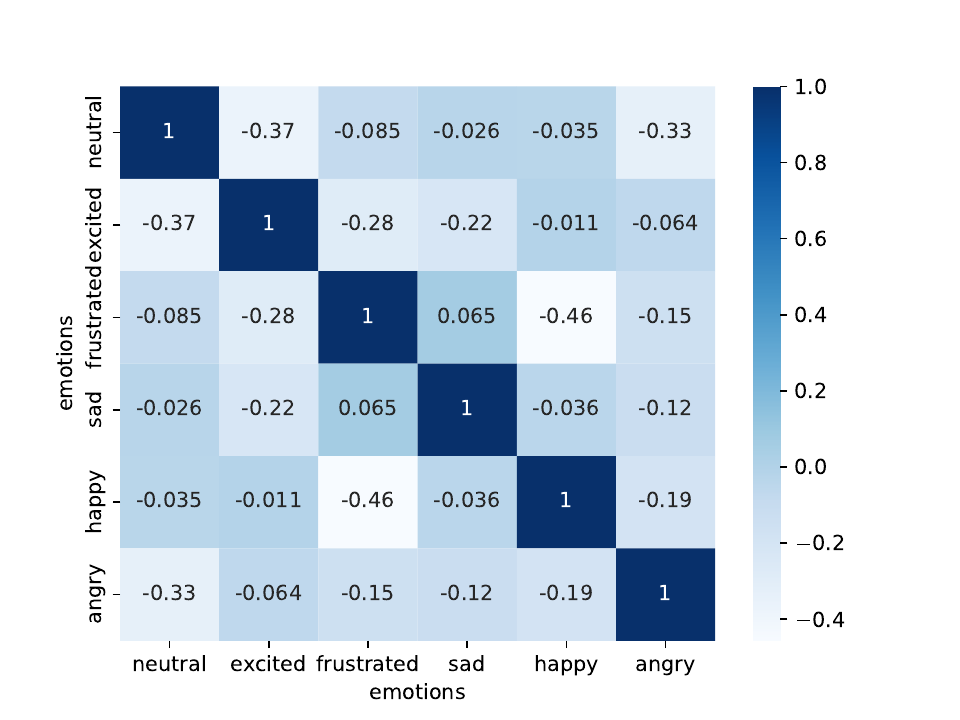}
    \caption{After training}
    \label{fig:after}
\end{subfigure}
\hfill
\begin{subfigure}{0.49\columnwidth}
    \centering
    \includegraphics[width=1.0\columnwidth]{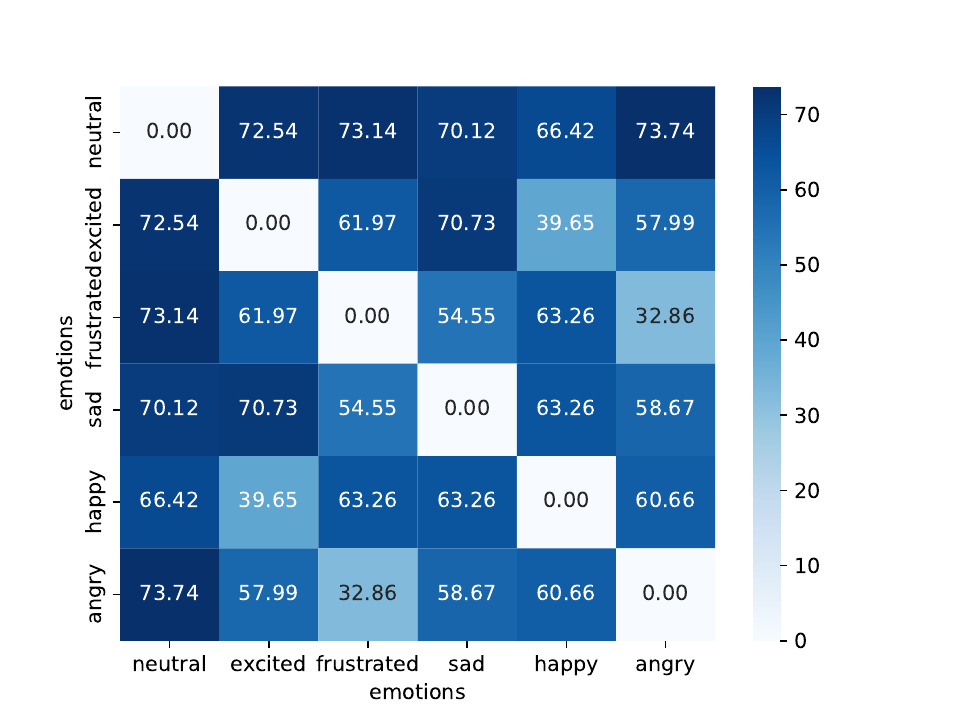}
    \caption{Before training}
    \label{fig:before angle}
\end{subfigure}
\hfill
\begin{subfigure}{0.49\columnwidth}
    \centering
    \includegraphics[width=1.0\columnwidth]{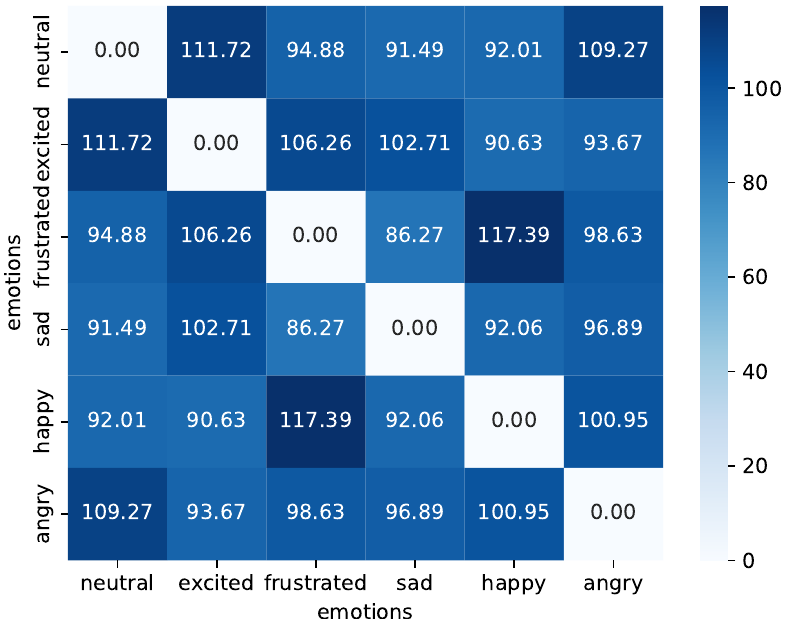}
    \caption{After training}
    \label{fig:after angle}
\end{subfigure}
\caption{The cosine similarity of pair-wise emotions. Figure (a) and (b) depicts cosine similarity between emotion anchors before and after training with EACL. (c) and (d) depicts the angle degree between emotion anchors before and after training with EACL respectively.}
\label{fig:before_and_after}
\end{figure}

\begin{figure}[t]
    \centering
    \includegraphics[width=0.85\columnwidth]{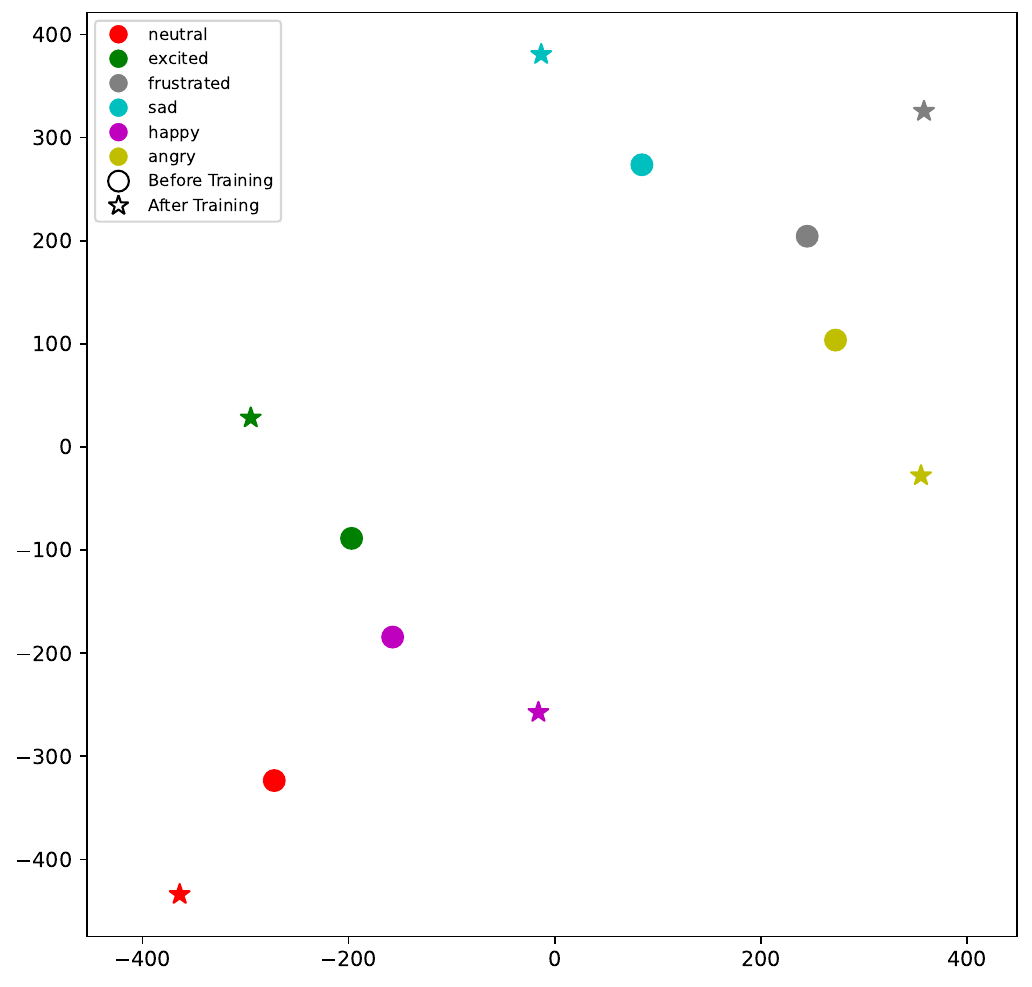}
    \caption{The t-SNE visualization of emotion anchors. Circles represent the position of emotion anchors before training and stars are the positions after training.}
    \label{fig:position}
\end{figure}

In this section, we conducted a comparison of the similarity between pairs of emotions before and after training with EACL in Figure~\ref{fig:before_and_after}. To observe the angle change more intuitively, we also include the angle degree. Figure~\ref{fig:before_and_after} reveals a significant decrease in similarity for emotion anchors that are considered similar. For instance, the cosine similarity between \textit{excited} and \textit{happy} drops sharply from 0.77 to 0.08, while for \textit{frustrated} and \textit{angry}, it decreases from 0.84 to -0.3. Meanwhile, naturally dissimilar emotions are now positioned further apart. 
For instance, the similarity between \textit{neutral} and other emotions also experiences a notable decline. These observations suggest that EACL effectively increases the separation between similar emotions, thereby enhancing the model's ability to distinguish between them. 
Figure~\ref{fig:position} visualizes the positions of anchors before and after training, where similar emotions are separated by EACL. 

\section{Conclusion and Future Work}
This paper introduces a novel framework for conversational emotion recognition called emotion-anchored contrastive learning. The proposed EACL leverages emotion representations as anchors to enhance the learning process of distinctive utterance representations. Building upon this foundation, we further adapt the emotion anchors through fine-tuning, bringing them the optimal positions and more suitable for classification purposes. Through extensive experiments and evaluations on three popular benchmark datasets, our approach achieves a new state-of-the-art performance. Ablation studies and evaluations confirm that the proposed EACL framework significantly benefits dialogue modeling and enhances the learning of utterance representations for more accurate emotion recognition.

The proposed EACL distributes the utterances in representation space more uniformly, which is beneficial for multi-class ERC tasks. When considering the context of multi-label classification, EACL can group relevant emotions guided by human knowledge, or adjust the inter-class weights of contrastive losses with label similarity~\cite{wang2022contrastive, zhao2022label}. Then, EACL can serve to detect multiple emotions in a single utterance, which will be left for future work.

\section*{Acknowledgement}
We would like to thank the anonymous reviewers for their valuable feedback. This work was supported by the NSFC (No. 62206126, 62376120, 61936012).

\section*{Limitations}
Our method focuses solely on textual inputs and does not incorporate multi-modal information. We recognize that complementing emotion recognition with facial expressions and tone can provide valuable information. Considering multi-modal inputs is an interesting direction for enhancements.

\section*{Ethics Statement}
The experiments conducted in this paper adopt open-source data for only research purposes. In this work, we try to facilitate machines with the ability to understand better human emotions which is beneficial for dialogue systems or robots. However, it is far from exceeding the understanding of humanity.

% Entries for the entire Anthology, followed by custom entries
\bibliography{anthology, custom}
\clearpage

\appendix
\section*{Appendix}
\section{Emotion Similarity Anlaysis}
\label{ESA}
\begin{figure}[h]
    \centering
    \includegraphics[width=0.95\columnwidth]{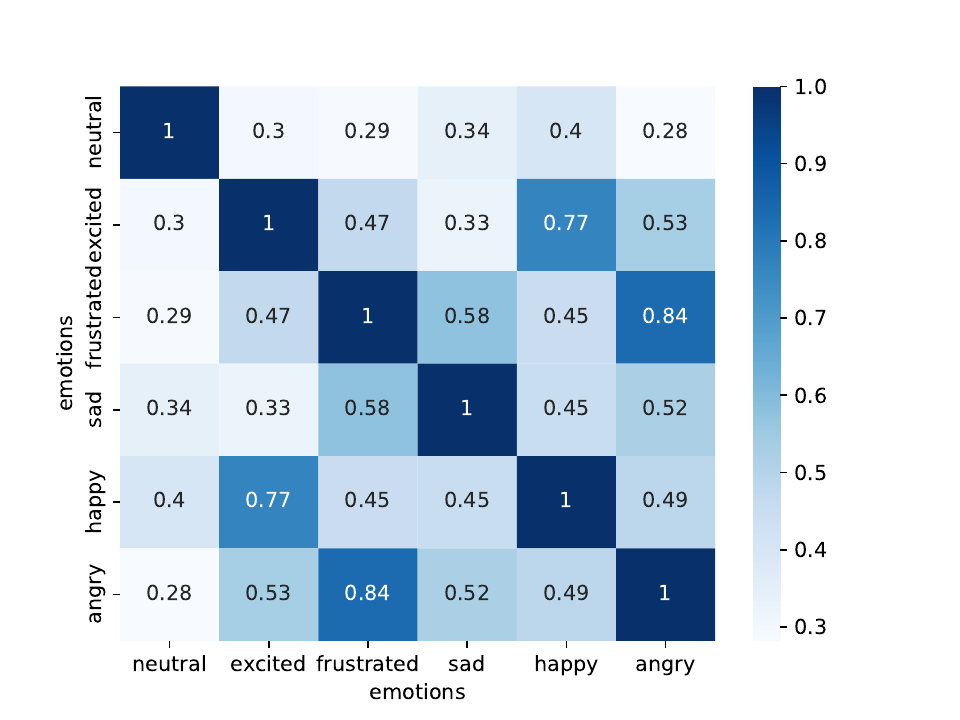}
    \caption{Cosine similarity between emotion word representations extracted from SimCSE-Roberta-Large~\cite{gao2021simcse}.}
    \label{fig:similarity}
\end{figure}

To better understand our motivation, we exhibit the emotion similarity in Figure \ref{fig:similarity}. We split the emotions into 3 groups which are composed of positive emotions, negative emotions, and neutral, where positive emotions include \textit{excited} and \textit{happy}, negative emotions contain \textit{frustrated}, \textit{sad}, \textit{angry}, and neutral. It is observed that \textit{excited} and \textit{happy} have a cosine similarity of 0.77, and for \textit{frustrated} and \textit{angry}, they have 0.84 cosine similarity. The similarity of the positive emotions group is higher than that of the negative emotions group. For \textit{neutral}, it is almost equally similar to other emotions.
\section{Experimental Setup}
\label{sec:hyper}
EACL loads the initial parameter by SimCSE-Roberta-Large~\cite{gao2021simcse} which is identical to the setting of SPCL. All the hyperparameters are reported in Table \ref{Hyperparameters}. We exploit grid-search for $\lambda_1$ in \{0, 0.1, 0.3, 0.5, 0.7, 0.9\}, $\lambda_2$ in \{0, 0.01, 0.1, 1.0\} and $\tau$ in \{ 0.05, 0.07, 0.1, 0.15, 0.2\}.
\begin{table}[H]
\resizebox{\columnwidth}{!}{%
\begin{tabular}{cccc}
\toprule

Hyperparameters       & IEMOCAP & MELD  & EmoryNLP \\ \midrule
$\lambda_1$           & 0.9     & 0.1   & 0.9     \\
$\lambda_2$           & 0.01    & 0.1   & 0.01     \\
Temperature $\tau$    & 0.1     & 0.1   & 0.15   \\ 
Epochs                & 8       & 8     & 8   \\ 
Maximum length        & 256     & 256   & 256   \\ 
Learning rate  & 1e-5    & 1e-5  & 1e-5   \\ 
Dropout               & 0.1     & 0.1   & 0.1   \\
\bottomrule
\end{tabular}%
}
\caption{Hyperparameters of EACL on three benchmark datasets.}
\label{Hyperparameters}
\end{table}

\begin{figure}[H]
\centering
\begin{subfigure}{0.49\columnwidth}
    \centering
    \includegraphics[width=1.0\linewidth]{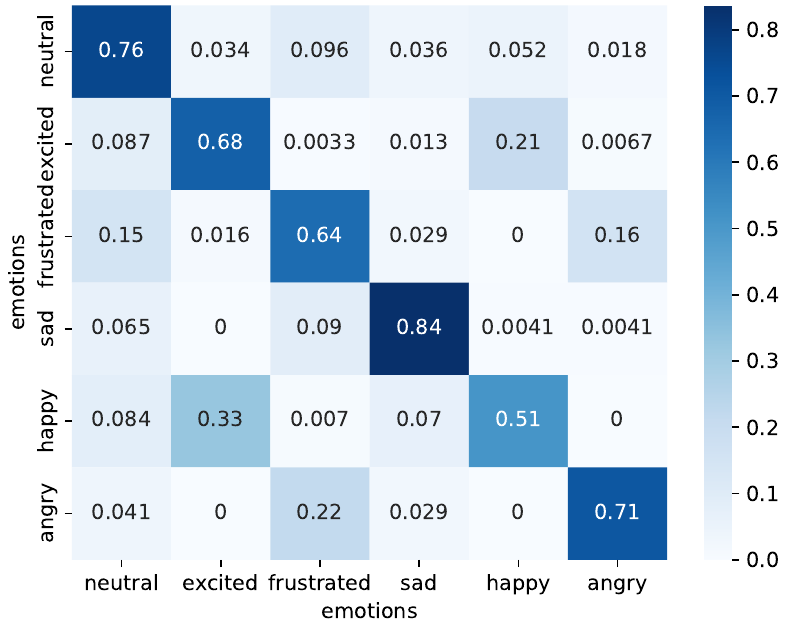}
    \caption{IEMOCAP (EACL)}
    \label{fig:iemo_EACL}
\end{subfigure}
\hfill
\begin{subfigure}{0.49\columnwidth}
    \centering
    \includegraphics[width=1.0\linewidth]{figures/cm_iemocap_spcl.pdf}
    \caption{IEMOCAP (SPCL+CL)}
    \label{fig:iemo_SPCL}
\end{subfigure}
\hfill
\begin{subfigure}{0.49\columnwidth}
    \centering
    \includegraphics[width=1.0\linewidth]{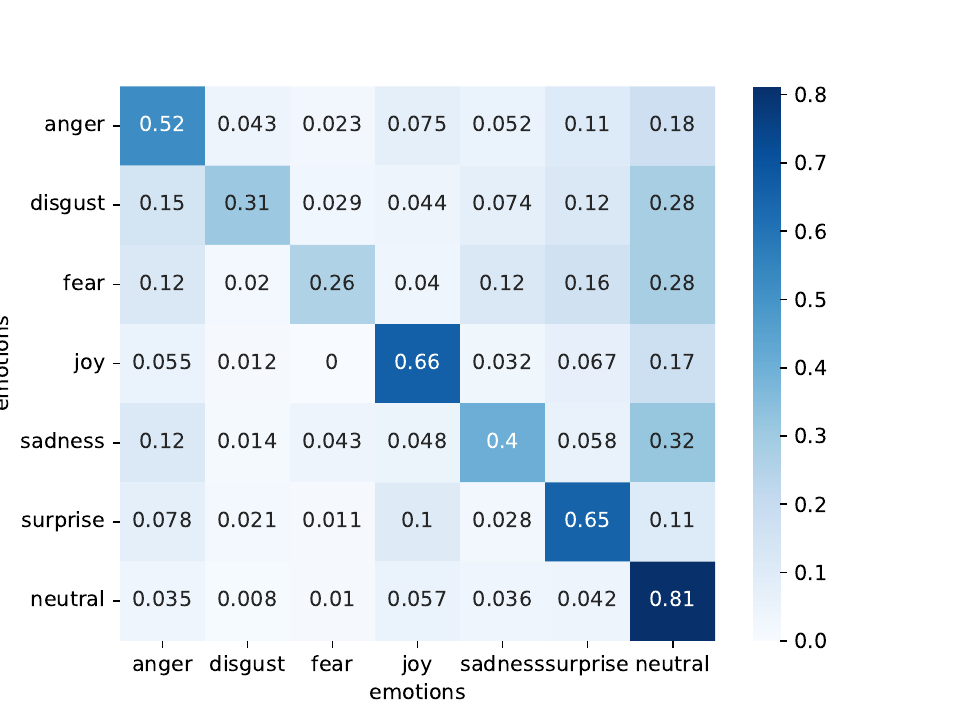}
    \caption{MELD (EACL)}
    \label{fig:meld_EACL}
\end{subfigure}
\hfill
\begin{subfigure}{0.49\columnwidth}
\centering
    \includegraphics[width=1.0\linewidth]{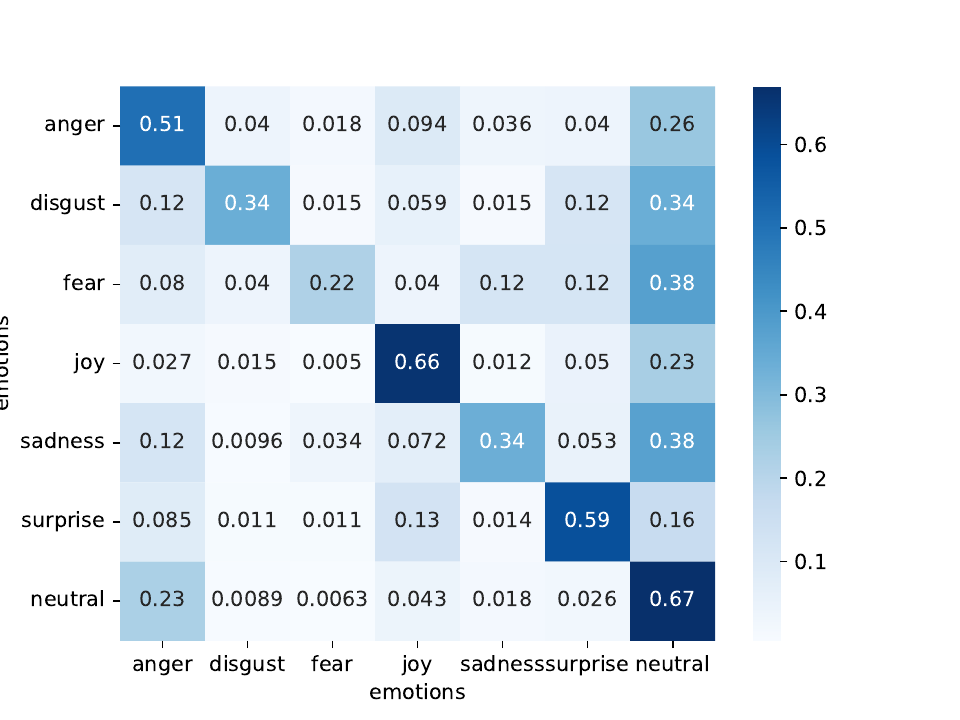}
    \caption{MELD (SPCL+CL)}
    \label{fig:meld_SPCL}
\end{subfigure}
\hfill
\begin{subfigure}{0.49\columnwidth}
\centering
    \includegraphics[width=1.0\linewidth]{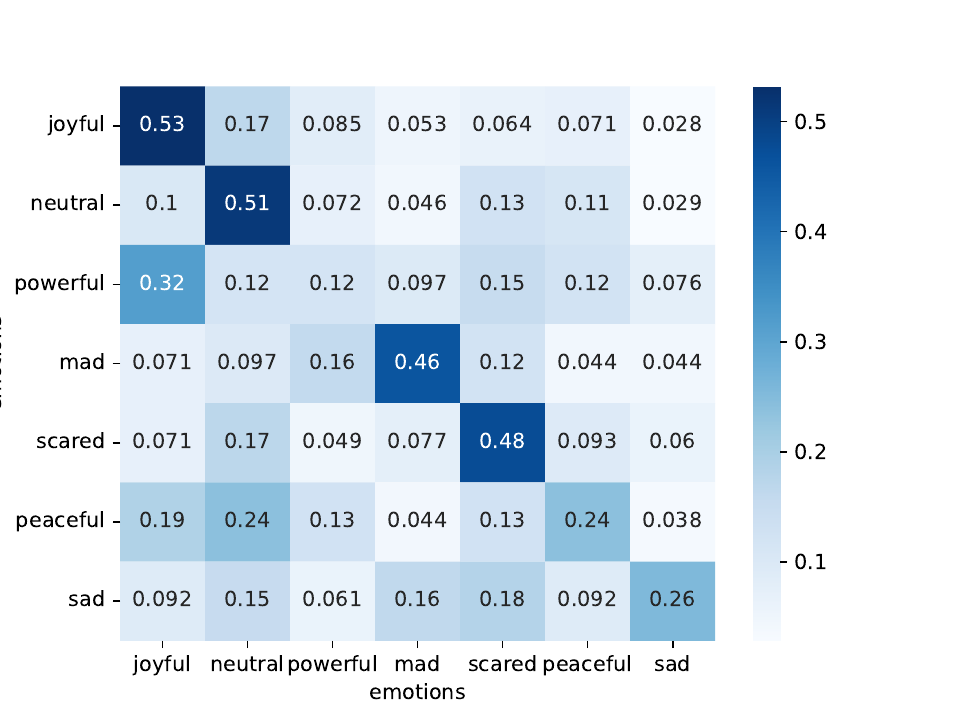}
    \caption{EmoryNLP (EACL)}
    \label{fig:emory_EACL}
\end{subfigure}
\hfill
\begin{subfigure}{0.49\columnwidth}
\centering
    \includegraphics[width=1.0\linewidth]{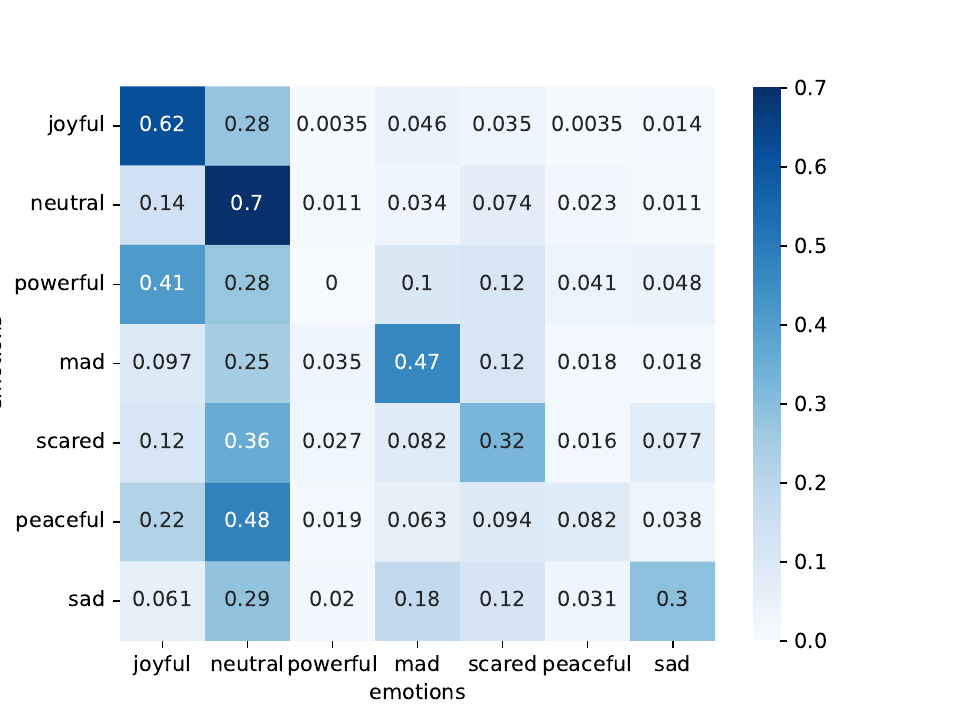}
    \caption{EmoryNLP (SPCL+CL)}
    \label{fig:emory_SPCL}
\end{subfigure}
        
\caption{The normalized confusion matrix of three benchmark datasets, each row is the true classes and column is predictions. The Coordinate $i, j$ means the percentage of emotion $i$ predicted to be emotion $j$.}
\label{fig:detailed}
\end{figure}
\section{Detailed Performance Analysis}
\label{DPA}
In Figure \ref{fig:detailed}, we provide the normalized confusion matrices for our EACL and SPCL+CL models across various datasets. These matrices serve as crucial tools for assessing the models' performance. Notably, when we examine the diagonal elements of these matrices, it becomes evident that EACL consistently outperforms the state-of-the-art method SPCL+CL in terms of true positives for most fine-grained emotion categories. This suggests that EACL excels at learning features that are more distinguishable.

Particularly noteworthy is the performance of EACL in comparison to SPCL+CL when considering specific emotion pairs, such as \textit{excited} and \textit{happy}, as well as \textit{frustrated} and \textit{angry} on the IEMOCAP dataset. In these cases, EACL demonstrates superior performance. This underscores the effectiveness of the EACL framework in effectively addressing the challenge of misclassification, especially when dealing with emotions that share similar characteristics.
When we focus on the MELD and EmoryNLP datasets, we observe that EACL significantly reduces misclassifications between \textit{neutral} emotions and other emotional states. This highlights EACL's capability to effectively mitigate misclassification issues not only for similar emotions but for all emotion categories.

\begin{table}[t]
\begin{subtable}{\linewidth}
    \caption{IEMOCAP}
    \resizebox{\columnwidth}{!}{%
        \centering
       \begin{tabular}{ccccccccc}
\toprule
Model       & Exc & Fru  & Sad & Neu & Ang & Hap & Avg & W-f1\\ \midrule
Deberta &     68.55&	69.74&	80.17&	70.18&	65.41&	50.96&	67.50&	69.09 \\
PromCSE & 	68.64&	67.19&	80.81&	74.66&	69.11&	53.41&	68.97&	70.45 \\ \midrule
SPCL+CL & 66.72  & 63.96  & 80.03  & 72.29  & 64.82  & 43.96  &  65.30  & 67.19 \\
\bottomrule
\end{tabular}%
}
\end{subtable}

\begin{subtable}{\linewidth}
\caption{MELD}
\resizebox{\columnwidth}{!}{%
    \centering
   \begin{tabular}{cccccccccc}
\toprule
Methods       & Fear & Neu  & Ang & Sad & Dis & Surp & Joy & Avg& W-f1\\ \midrule
Deberta &    34.0&	80.43&	55.28&	44.44&	37.59&	60.85	&65.34&	53.99&	67.8
 \\ 
PromCSE & 	23.59&	81.0	&54.96&	43.35&	30.53&	59.51	&65.12&	51.15&	67.38
\\\midrule
SPCL+CL    & 26.59  & 77.92  &  54.40  &  43.53 & 30.94 &  59.26 & 60.34  & 50.43  &  65.74\\  \bottomrule
\end{tabular}%
}
\end{subtable}
\begin{subtable}{\linewidth}
\caption{EmoryNLP}
\resizebox{\columnwidth}{!}{%
\centering
\begin{tabular}{cccccccccc}
\toprule
Methods       & Joy & Sad  & Pow & Mad & Neu & Pea & Sca & Avg & W-f1\\ \midrule
Deberta &    54.04	&28.74&	21.54	&41.73&	51.75&	18.12&	42.52&	36.92&	41.09\\ 
PromCSE & 	54.42&	28.33&   14.21	&43.35&	51.64&	23.42	&41.30	&36.68	&40.93 \\ \midrule
SPCL+CL& 53.52 & 31.61  & 10.28 & 44.21 & 51.40 & 16.83& 39.51 & 35.34 & 39.52\\  \bottomrule
\end{tabular}%  
}

\end{subtable}
\caption{Fine-grained performance record on different language models for all emotions on three benchmark datasets, the F1-score is used for each class. }
\label{fine-diff}
\end{table}
\section{Fine-Grained Performance on Different Models}
\label{sec:fine-diff}
In this section, we report the fine-grained performance when using Deberta-Large~\cite{he2020deberta} and Promcse-Roberta-Large~\cite{jiang2022improved} in Table~\ref{fine-diff}. 
The results indicate that our learning framework is robust to different language models. Similar to the result under Roberta-SimCSE, these models can also effectively separate similar emotions and achieve state-of-the-art performance on the benchmark datasets.

\end{document}